\pdfoutput=1

\documentclass[11pt]{article}

\usepackage[final]{coling}

\usepackage{times}
\usepackage{latexsym}

\usepackage[T1]{fontenc}

\usepackage[utf8]{inputenc}

\usepackage{microtype}

\usepackage{inconsolata}

\usepackage{graphicx}
\usepackage{amsmath}
\usepackage{amssymb}
\usepackage{makecell}
\usepackage{multirow}
\usepackage{threeparttable}
\usepackage{pifont}
\usepackage{ulem}
\usepackage{subcaption}

\title{
  \begin{minipage}{1cm} 
    \vspace{-0.5em}
    {\includegraphics[width=1.2cm]{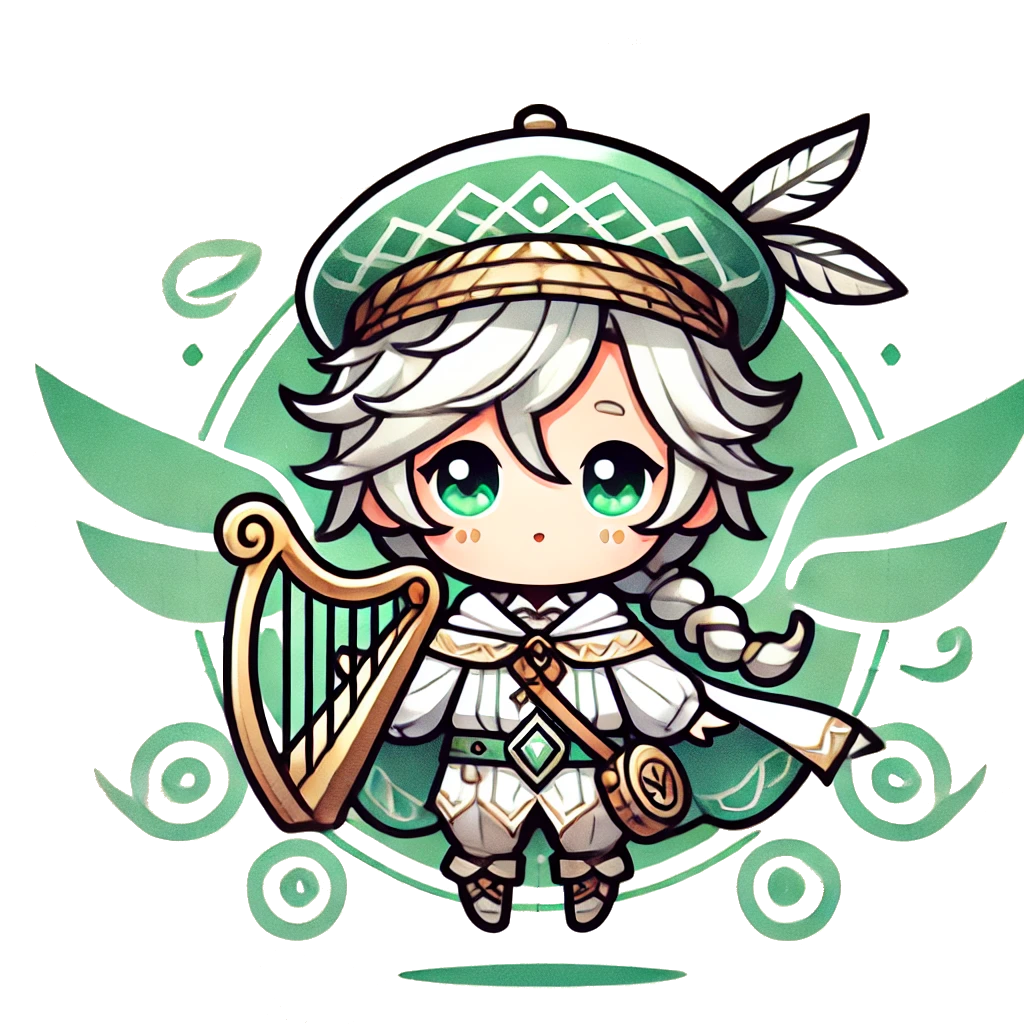}} 
  \end{minipage}%
  \hspace{0.2cm} 
  \begin{minipage}{0.8\textwidth} 
    \centering
    \textbf{Minstrel: Structural Prompt Generation with Multi-Agents Coordination for Non-AI Experts} 
  \end{minipage}
}

\author{
 \textbf{Ming Wang\textsuperscript{1}},
 \textbf{Yuanzhong Liu\textsuperscript{2}},
 \textbf{Xiaoyu Liang\textsuperscript{3}},
 \textbf{Yijie Huang\textsuperscript{1}},
\\
 \textbf{Daling Wang\textsuperscript{1}},
 \textbf{Xiaocui Yang\textsuperscript{1}},
 \textbf{Sijia Shen\textsuperscript{4}},
 \textbf{Shi Feng \textsuperscript{1}},
\\
 \textbf{Xiaoming Zhang\textsuperscript{1}},
 \textbf{Chaofeng Guan\textsuperscript{4}},
 \textbf{Yifei Zhang\textsuperscript{1}},
\\
\\
 \textsuperscript{1}Northeastern University,
 \textsuperscript{2}Wuhan University,
 \textsuperscript{3}Zhejiang University,
 \textsuperscript{4}Shenyang Aerospace University
\\
 \small{
   \textbf{Correspondence:} \href{mailto:wangdaling@cse.neu.edu.cn}{wangdaling@cse.neu.edu.cn}
 }
}

\begin{document}
\maketitle
\begin{abstract}
LLMs have demonstrated commendable performance across diverse domains. Nevertheless, formulating high-quality prompts to assist them in their work poses a challenge for non-AI experts. Existing research in prompt engineering suggests somewhat scattered optimization principles and designs empirically dependent prompt optimizers. Unfortunately, these endeavors lack a structural design, incurring high learning costs and it is not conducive to the iterative updating of prompts, especially for non-AI experts. Inspired by structured reusable programming languages, we propose LangGPT, a structural prompt design framework. Furthermore, we introduce Minstrel, a multi-generative agent system with reflection to automate the generation of structural prompts. Experiments and the case study illustrate that structural prompts generated by Minstrel or written manually significantly enhance the performance of LLMs. Furthermore, we analyze the ease of use of structural prompts through a user survey in our online community\footnote{\href{https://langgpt.ai}{https://langgpt.ai}}.
\end{abstract}

\section{Introduction}
LLMs make it possible to instruct computers in natural language, attracting the attention of many non-AI experts, who have shown keen interest in using LLMs for code generation \citep{nejjar_llms_2024}, analyzing scientific literature \citep{cai_sciassess_2024}, discovering scientific equation \citep{shojaee_llm-sr_2024}, etc. Also, they enslave LLMs to execute many specific scientific tasks, like predicting properties of molecules and materials \citep{jablonka_14_2023}, simulating biological systems \citep{schaefer_large_2023}, and stellar lightcurve classification \citep{li_deep_2024}. The widespread adoption and recognition of LLMs have significantly motivated practitioners and researchers in related fields. However, despite the impressive performance of LLMs on diverse tasks \citep{qin_is_2023,jiao_is_2023} across varied domains \citep{li2023chatdoctor,wu_bloomberggpt_2023,zhang_affective_2024}, fully harnessing their capabilities remains a challenge \citep{eric_complete_2022,chen_unleashing_2023,nagaraju_gajula_guide_2023}. Therefore, prompt engineering, an empirical science dedicated to effectively communicating with and eliciting desired outputs from LLMs, has been widely focused on \citep{tanay_varshney_introduction_2023,mesko_prompt_2023,zhenxuan_wang_how_2023}.
\begin{figure}[ht]
    \centering
    \includegraphics[width=\linewidth]{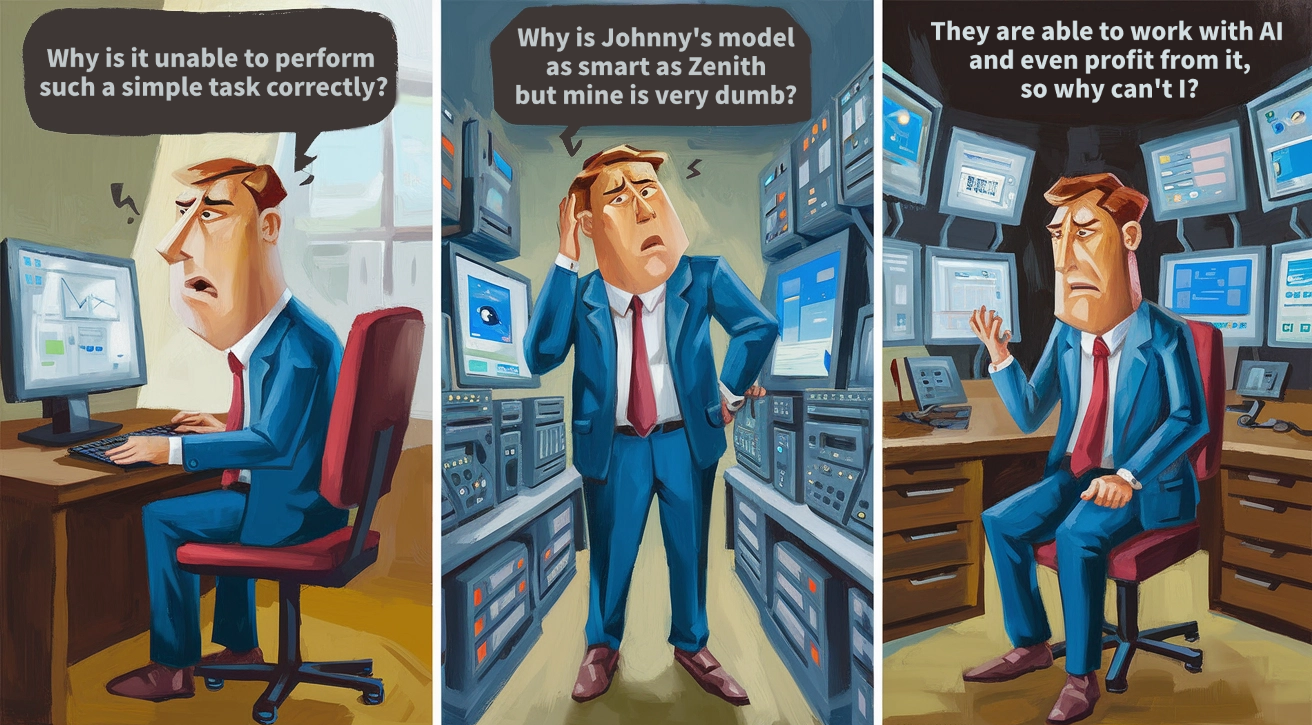}
    \caption{Complaints voiced by Oliver when using LLMs. The name `Oliver' is fictional, and `Johnny' comes from \citep{10.1145/3544548.3581388}.}
    \label{fig:background}
\end{figure}

For a non-AI expert, it can be frustrating to leverage Large Language Models (LLMs), which appear complex, hard to master, and frequently demand technical skills or costly resources. As exemplified in Fig. \ref{fig:background}, Oliver will struggle with his inability to use LLMs effectively, feel envious of Johnny's mastery of them, and even worry about falling behind the times.
To make it easier for non-AI experts to master LLMs, the researchers optimize prompts mainly by summarising design tricks \citep{li_dialogue_2023} and designing automatic optimization methods \citep{sun_autohint_2023}. Summarising design tricks extracts tips for writing prompts by analyzing a large number of prompts that lead to differences in the corresponding outputs. However, there are very large differences in the sensitivity of different models to prompts, and different tasks are affected by prompts to very different degrees. Automatic prompt optimization methods, on the other hand, usually require adjusting prompts based on incorrect responses. Thus, these methods are usually only applicable to tasks for which there are reference answers.

Inspiring by the belief that prompt is the programming language of the LLM era \citep{10.1145/3626252.3630909,mund_ai_2023}, we propose \textbf{Lang}uage for \textbf{GPT}-like LLMs (LangGPT), a structural prompt framework. LangGPT combines the systematic, prescriptive, and reusable characteristics of programming language with the flexibility and extensibility of natural language. Drawing inspiration from programming language, we design LangGPT with a dual-layer structure composed of modules and elements. Modules represent the perspectives of the content.
In addition, we define elements as specific content within modules and provide guidelines for writing them. 

Further, we propose Minstrel, a structural prompt generation tool based on a \uline{m}ult\uline{i}-age\uline{n}t \uline{s}ys\uline{t}em with \uline{re}f\uline{l}ection. Minstrel generates structural LangGPT prompts automatically employing multiple agents' collaboration. Minstrel consists of three working groups: the analysis group, the design group, and the test group. The analysis group first analyses the task proposed by users and activates the corresponding module agents in the design group. Moreover, the test group systematically tests the effectiveness \citep{openai_six_nodate} of the prompts and identifies their shortcomings using the LLM-based reflector agent in the analysis group, which the corresponding module agents then optimize.
Experiments and the case study demonstrate that structural prompts Minstrel-generated or hand-designed are better than baseline prompts for bootstrapping LLMs. In addition, we conduct user research to prove the ease of use and user satisfaction of structural prompts in our prompt community (see Fig. \ref{fig:community}). In addition, we invite ordinary users to design prompts to further validate the ease of use of structural prompts and verify the gain effect of structural prompts on different performance models.

In summary, the contributions of this work include: (1) We propose a structural prompt design framework LangGPT to improve the generalization and reusability of prompts, reducing the learning cost of prompt design. The code is in \href{https://github.com/langgptai/LangGPT}{https://github.com/langgptai/LangGPT}. (2) We propose Minstrel, a generation tool to design structural LangGPT prompts through multi-agent collaboration. The code is in \href{https://github.com/sci-m-wang/Minstrel}{https://github.com/sci-m-wang/Minstrel}. (3) We demonstrate through experiments and case studies that structural prompts (Minstrel-generated and manually) can better guide LLMs. Meanwhile, we organize a user survey to analyze the effectiveness and ease of use of structural prompts based on the prompt community we built.

\section{Related Works}
\subsection{Prompt Design Tricks}
Prompt engineering mainly involves designing and optimizing the prompts. Some researchers first explore some tricks for prompt optimization. \citet{bsharat_principled_2023} introduce 26 guiding principles designed to improve the performance of LLMs. \citet{bjerg_tips_2024} collects tips for prompt optimization and provides exercises from basic, advanced, and contextual perspectives. \citet{liu_design_2022} evaluate 5493 generations covering 51 themes and 51 styles throughout five experiments in the text-to-image task and summaries prompt design guidelines. The absence of a systemic design to these tricks makes it challenging to memorize, understand, and master them fully. More concerning, these methods are often closely tied to specific models and tasks, resulting in limited generalizability and robustness. For instance, \citet{bsharat_principled_2023} believe that there is no need to be polite to LLMs, while \citet{yin_should_2024} believe that treating LLMs with proper respect reduces bias and improves performance.

\subsection{Automatic Prompt Optimization}
In addition to these tricks, some researchers also focus on optimizing prompts automatically. \citet{li_dialogue_2023} design a multi-round dialogue alignment strategy and utilize GPT-4 \citep{achiam_gpt-4_2023} to generate a readability prompt set. Meanwhile, they propose an efficient prompt screening metric that can filtrate high-quality prompts with linear complexity.  \citet{sun_autohint_2023} guide LLMs to derive new prompts for a given instance from the incorrect reasoning, then summarise the corresponding prompts for each instance as a reference for optimizing prompts. \citet{pryzant_automatic_2023} leverage mini-batches of data to form natural language ``gradients'' and utilize beam search and bandit selection procedure to edit the current prompt in the opposite semantic direction of the gradient. \citet{fan_towards_2023} analyze a large prompt database and present an automatic prompt optimization framework. \citet{wang_promptagent_2023} introduce PromptAgent which can reflect on model errors and generate constructive error feedback to induce precise expert-level insights and in-depth instructions. \citet{guo_connecting_2023} connect LLMs with evolutionary algorithms and proposes a novel framework for discrete prompt optimization, called EvoPrompt. \citet{hao_optimizing_2022} and \citet{cheng_black-box_2023} optimize prompts by aligning human and LLMs' preference styles. \citet{10.1145/3544548.3581388} take a prototype LLM-based chatbot design tool as the design probe, supporting non-AI-experts engage in ``end-user prompt engineering''. These automated methods for optimizing prompts often depend on standard labeling or manual intervention to correct erroneous outputs. Thus, they are usually not well adapted to open-generation tasks.

\subsection{Prompt Design Framework}
Prompt optimization can significantly improve the performance of LLMs. Nevertheless, neither prompt tricks nor automatic prompt optimization methods are systematically designed, resulting in high learning costs and it is not conducive to iterative updating of prompts. Moreover, they are usually only applicable to certain models or tasks, with poor generalization and robustness. Consequently, some researchers consider constructing complete prompt design frameworks. \citet{nigh_chatgpt3-free-prompt-list_nodate} collects vast quality prompts and summaries of the CRISPE framework. COSTAR template, which consists of context, objective, style, tone, audience, and response, is introduced in \citep{govtech2023prompt}.

However, these frameworks are rigid and they only outline the necessary aspects of prompts without offering specific guidelines for crafting each one. Most critically, they fail to provide prompt generation tools to support the theory, leaving users to rely on manual writing based on their own experience. As a result, they lack generalization and scalability, still requiring significant effort and costs for non-AI experts to master.

\section{LangGPT: Structural Prompt Design Framework}
As programming language is more systematic and reusable, we first analyze the differences between natural languages and programming languages to design a generalizable, reusable, and extensible prompt framework that is easy to master for non-AI experts. While natural language is primarily used for communication, the programming language is designed to instruct machines to execute tasks \citep{sebesta_concepts_2012}. The main difference between the two languages is that natural language is more vague and flexible, while programming language is more standardized and precise \citep{chakray_programming_2018,fromkin2018introduction}.
Drawing on these characteristics, we propose LangGPT, a structural prompt framework, combining the advantages of natural language and programming language.

\subsection{Overall Dual-layer Structure}
To systematically design prompts that meet the principles, we have made full reference to the design ideas and structures of object-oriented programming languages \cite{rentsch_object_1982,lutz_programming_2010}.
We refer to the structure of the programming language propose a dual-layer structure for prompt design, and define modules and elements for prompts.
Modules are similar to classes in programming language, and each module represents an aspect of the requirements for LLMs. 
Within a module, many internal elements are included. Elements are similar to functions and properties in programming languages and represent the content of direct and specific instructions to LLMs. For example, ``Output should be no more than 500 words'' could be an element in a prompt that belongs to the module \textbf{constraint}.
Furthermore, we design basic elements for different modules.


\begin{table*}[!htb]
    \centering
    \begin{tabular}{c|l}
    \hline
    \textbf{Module}                    & \textbf{Samples of Basic Elements}                                               \\ \hline
    Role                      & Magazine Editor \\ \hline
    Profile                   & $\bullet$ Language: English                                     \\ \hline
    Goal                      & $\bullet$ You need to generate a title for the article.                           \\ \hline
    Constraint                & $\bullet$ The length of the title should not exceed 20 words.                     \\ \hline
    \multirow{6}{*}{Workflow} & \#\#\# Extracting the kernel content \\
                              & $\bullet$ For the given article $\langle \textsc{article} \rangle$, please execute the following actions:         \\
                              & \quad $\circ$ Analyse the theme of the article;                                       \\
                              & \quad $\circ$ Detecting the main objects and related things described in the article; \\
                              & \quad $\circ$ Summarising the core content from the article;                          \\
                              & \quad $\circ$ Save the kernel content.                                                \\ \hline
    Style                     & $\bullet$ The style of the title should be formal.                                \\ \hline
    \end{tabular}
    \caption{Examples of basic internal elements of inherent modules in the writing scenario. This prompt leads LLMs to generate a title for a given article. We have chosen five modules as examples i.e. \textbf{profile}, \textbf{goal}, \textbf{constraint}, \textbf{workflow}, and \textbf{style}. Particularly, for the workflow module, we show a function-like element.}
    \label{tab:basic_elements}
\end{table*}

\subsection{Modules}
Building on prior research, design of generative agents \cite{park_generative_2023}, and practical experience from our community, we define basic modules for prompt construction.

\textit{Role} is the name of prompts and also the role set for LLMs. \textit{Profile} is designed to facilitate the version management and updating of prompts, containing some basic information about the prompts, such as author, version, description, language, etc. \textit{Constraints}, which some users prefer to call ``Attention'', i.e., the prohibitions that cannot be violated and requirements that must be met when generating responses. \textit{Goals} lists the final objective that the user wants to achieve, which is what the LLMs need to accomplish. \textit{Initialization} is a flag that informs LLMs that they are about to start a dialogue. Sometimes a specified first sentence is also given in this module. \textit{Examples} gives LLMs input-output pairs as instances to learn from. Similar to the Chain of Thoughts approach \cite{wei_chain--thought_2023}, users can instruct LLMs on the process when executing a task in the \textit{Workflow} module. It is often necessary to instantiate this module when the task requirements are more complex. \textit{Skills} activates the task-relevant segment of the massive knowledge and myriad capabilities that LLMs gain during training. \textit{Suggestion} provides advice and behavioral planning for LLMs in branching situations when performing tasks. This module lists common scenarios and gives behaviors or responses that LLMs can take in such situations. \textit{Background} indicates the contextual information and the memories that LLMs must have when performing their tasks. \textit{Style} qualifies the style, tone, and affective characteristics of LLMs' responses. \textit{Output format} specifies the format of LLMs' responses. Specifying the output format improves the efficiency and accuracy of the results extraction in certain tasks. To keep LLMs-based assistants from being stuck with a single action, we add the \textit{Command} module. For example, in the ``Who is The Spy'' game \footnote{\href{https://github.com/sci-m-wang/Spy-Game}{https://github.com/sci-m-wang/Spy-Game}} using LangGPT designed by a user in our community, LLM players can either describe their keywords or vote rival players out of the game.

\subsection{Elements}

In addition to understanding which aspects of designing the prompt from, how to write the exact content of prompts can be challenging for non-AI experts. Therefore, we analyze the composition of prompts and design the elements that make up the prompts, relying on cases of quality prompts shared by users and community exchange content. Three purposes are typically included in prompts: (1) Implying a certain info to LLMs; (2) Letting LLMs execute a certain task with or without output; (3) The combination of the first two.

\begin{figure*}[!ht]
    \includegraphics[width=\textwidth]{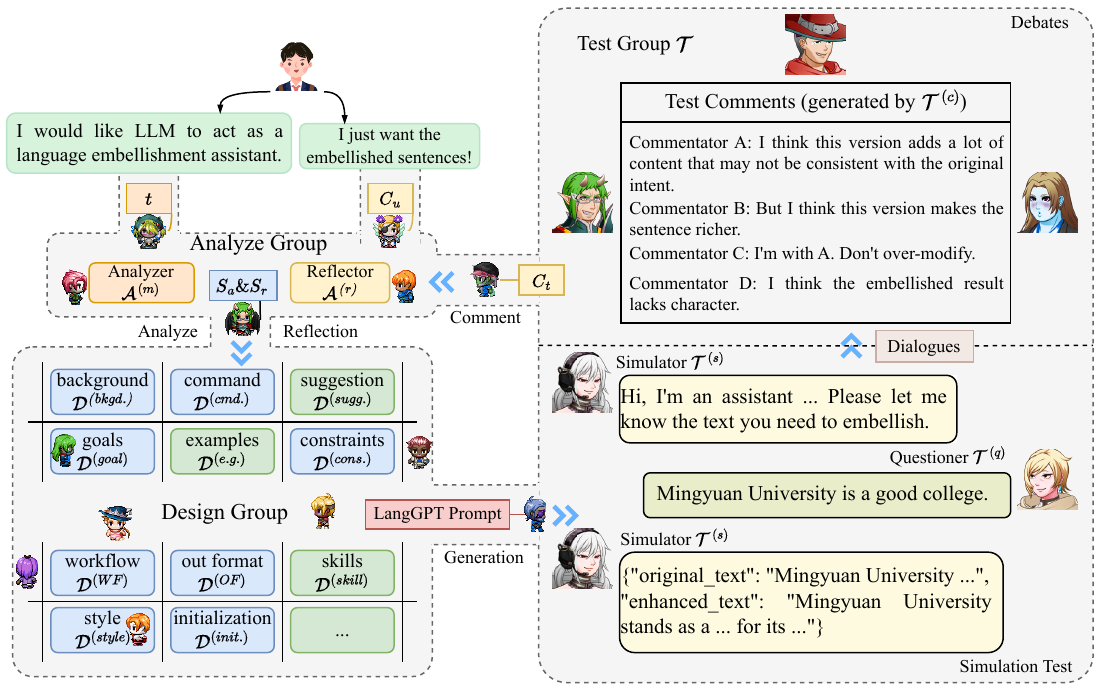}
    \caption{The overall framework of Minstrel, a structural prompt generation framework with multi-agents collaboration. There are three working groups: \textit{analyze group}, \textit{design group}, and \textit{test group}. In \textit{design group}, \textcolor[HTML]{A3BBDD}{blue modules} indicate activated modules, and \textcolor[HTML]{ABCD9D}{green} indicates modules that are not required for the current task and are not activated.}
    \label{fig:overall_framework}
\end{figure*}

The first of these is very similar to the assignment of properties or variables in programming languages. Correspondingly, the last two categories are similar to functions in programming languages. Thus, we construct these three types of basic elements. We use designed templates like ``The $\langle \textsc{property} \rangle$ is $\langle \textsc{value} \rangle$.'' to simulate an assignment. For the latter two cases, it is necessary to specify the input information, the task, and the output, where input and output can be omitted. We simulate functions using a form like this: ``For the given $\langle \textsc{property} \rangle$ of $\langle \textsc{value} \rangle$, please execute the following actions: $\langle \textsc{actions} \rangle$; Return the $\langle \textsc{result} \rangle$.'' In the basic element writing patterns we provide, the contents contained in the \textbf{angle brackets} need to be populated according to the module and the usage scenario. It is important to note that the writing patterns we provide only specify the idea of writing internal elements. To improve the generalisability and flexibility of prompts, the language can be adapted to express key information. In Table \ref{tab:basic_elements} we show some examples of the basic elements in some modules.

\section{Minstrel: Structural Prompt Generation with Multi-Agents Coordination}
To further reduce the learning cost of the LangGPT framework, we propose Minstrel, generating structural prompts through the collaboration of multiple generative agents. Minstrel introduces \textbf{working groups}, which are like different departments in a software development studio, collaborating but with a division of labor. Based on the working group, structural prompt generation is divided into low-coupled and flexible sub-tasks.  The overall framework is shown in Fig. \ref{fig:overall_framework}. Under the control of Unified Analysis, agents corresponding to different modules work together to generate structural prompts. In addition, Minstrel systematically tests and further optimizes the prompts by means of multi-agent debates.

\subsection{Working Group}
Although LLMs have a strong understanding of natural language and can assist in optimizing prompts, it remains challenging to fully grasp the task, break down the process, optimize the representation, and translate the user's requirements into a structural prompt that defines an assistant. Minstrel thus splits the prompt design into different working groups, with agents within each group responsible for being in charge of similar atomic tasks. Atomic tasks refer to tasks that are performed by a single agent and do not require further splitting. Specifically, there are three working groups in Minstrel: the Analysis Group ($ \mathcal{A} $), the Design Group ($\mathcal{D}$), and the Test Group ($\mathcal{T}$). $ \mathcal{A} $ analyzes the user requirements, feedback, and test results. $\mathcal{D}$ is responsible for the main generation of different modules. $\mathcal{T}$ performs systematic testing of the generated structured prompts.

\subsection{Agents in Working Groups}
There are analyzer $\mathcal{A}^{(m)}$ and reflector $\mathcal{A}^{(r)}$ in $ \mathcal{A} $. $ \mathcal{A}^{(m)}$ analyses the task requirements from the user and activates the corresponding module design agents. $ \mathcal{A}^{(r)}$ summarises the issues from the test feedback and the user feedback and reflects them to the corresponding module design agents. $ \mathcal{D}$ consists of agents for designing different modules of prompts. These agents correspond to the modules in LangGPT, i.e. there is a specialized agent $\mathcal{D}^{(cons.)}$ for generating the content of constraints and a specialized agent $\mathcal{D}^{(bkgd.)}$ for generating the content of backgrounds, and the same goes for the other modules. Simulator $\mathcal{T}^{(s)}$, questioner $\mathcal{T}^{(q)}$ and commentators $\mathcal{T}^{c}$ are included in $\mathcal{T}$. $\mathcal{T}^{(s)}$ followed the prompts generated by $ \mathcal{D}$ to act as an assistant in dialogue with $\mathcal{T}^{(q)}$. Commentators $\mathcal{T}^{(c)}$ comment on the performance of the $\mathcal{T}^{(s)}$ and they can debate with each other.

\subsection{Multi-Agents Collaboration with Reflection}

For a task $t$, the entire design pattern involves both design (shown in Equ. (\ref{equ:design})) and reflection (shown in Equ. (\ref{equ:reflection})).

\begin{equation}
    \begin{aligned}
        S_a = \mathcal{A}^{m}(t), \hspace{2.9cm} \\
        M[k] = \mathcal{D}^{(k)}(t)~\text{for}~k~\text{in}~S_a,  \hspace{0.7cm}
    \end{aligned}
    \label{equ:design}
\end{equation}
where $S_a$ denotes the module activation state obtained by $ \mathcal{A}^{(m)}$, $M$ denotes the list storing the contents of the modules in LangGPT generated by $ \mathcal{D}$, and $\mathcal{D}^{(\cdot)}$ denotes the designer of the corresponding activated module.

\begin{equation}
    \begin{aligned}
        C_t = \mathcal{T}^{(c)}(\mathcal{T}^{(s)}(\sum M),~\mathcal{T}^{(q)}(t)), \\
        S_r = \mathcal{A}^{(r)}(C_t~+~C_u), \hspace{1.9cm} \\
        M[k] = M[k]~+~\mathcal{A}^{m}(S_r), \hspace{1.3cm}
    \end{aligned}
    \label{equ:reflection}
\end{equation}
where $C_t$ denotes the comments derived by $\mathcal{T}$, $C_u$ denotes the user's comments, and $S_r$ denotes the module activation state generated by $\mathcal{A}^{(r)}$ based on the comments. The modifications required for each activated module are summarised in $S_r$. Ultimately, the content of each module in $M$ is combined to obtain the final structural prompt.

When the user proposes a task that needs to be performed by LLMs, the task is first analyzed by $\mathcal{A}^{m}$, and the required modules in the design group are activated. Then, the agents corresponding to the activation modules in $\mathcal{D}$ generate the contents that should be included in the prompt. Minstrel now generates the first version of the structural prompt. Testing and reflection are also needed to get better-quality prompts. $\mathcal{T}^{(s)}$ receives the prompt and acts as an assistant defined by the prompt. $\mathcal{T}^{(q)}$ is adaptive to the task communicates with the $\mathcal{T}^{(s)}$ to produce test dialogues. Commentators in $\mathcal{T}$ will evaluate the assistant's performance based on the test dialogues. These commentators were presupposed to have weak positions, two tending to criticize, two tending to favor, and one remaining neutral. They will evaluate the test dialogues according to their position. $\mathcal{A}^{r}$ will activate the corresponding modules based on the commentators' comments and the user's comments, prompting them to modify the corresponding content.

\section{Experiments}
We compared the effects of hand-designed and Minstrel-generated LangGPT prompts with those designed by the baseline methods on the performance of bootstrapped LLMs.

\subsection{Experiment Settings}

\subsubsection{Large Language Models}
\label{llms}
The LLM used in Minstrel is GPT-4-turbo \cite{achiam_gpt-4_2023}. To evaluate the effectiveness of the prompts, we use Gemma-2-9b-it \cite{team_gemma_2024}, Qwen2-7B-Instruct \cite{bai_qwen_2023}, Meta-Llama-3.1-8B-Instruct \cite{llama3modelcard}, Mistral-7B-Instruct-v0.2 \cite{noauthor_mistral-7b-instruct-v02_nodate}, GPT-4o-mini and Claude3-haiku \citep{anthropic_claude_nodate} to act as assistants defined by structural prompts.

\begin{figure*}[!h]
    \centering
    \begin{subfigure}[b]{0.24\textwidth}
        \includegraphics[width=\textwidth]{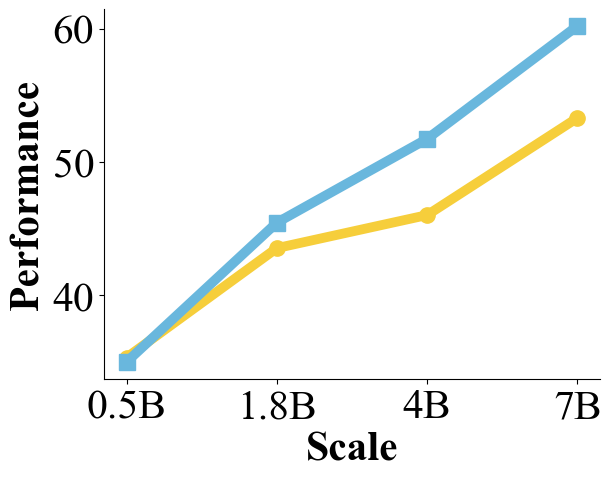}
        \caption{Average}
        \label{fig:average}
    \end{subfigure}
    \begin{subfigure}[b]{0.24\textwidth}
        \includegraphics[width=\textwidth]{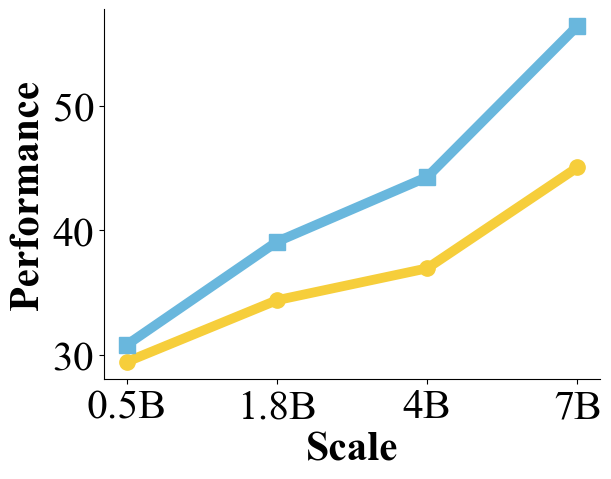}
        \caption{ARC-Challenge}
        \label{fig:arc}
    \end{subfigure}
    \begin{subfigure}[b]{0.24\textwidth}
        \includegraphics[width=\textwidth]{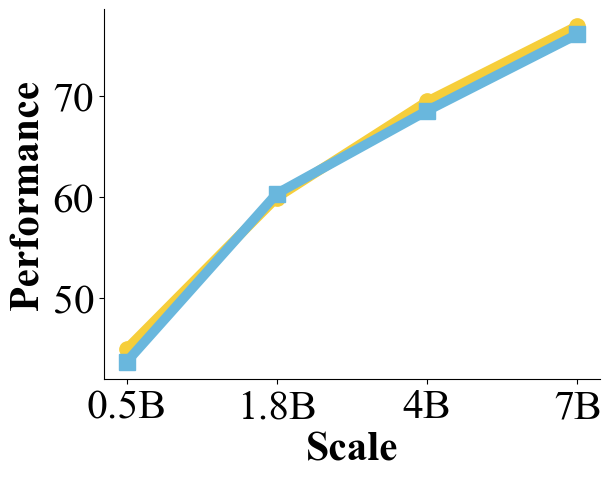}
        \caption{Hellaswag}
        \label{fig:hellaswag}
    \end{subfigure}
    \begin{subfigure}[b]{0.24\textwidth}
        \includegraphics[width=\textwidth]{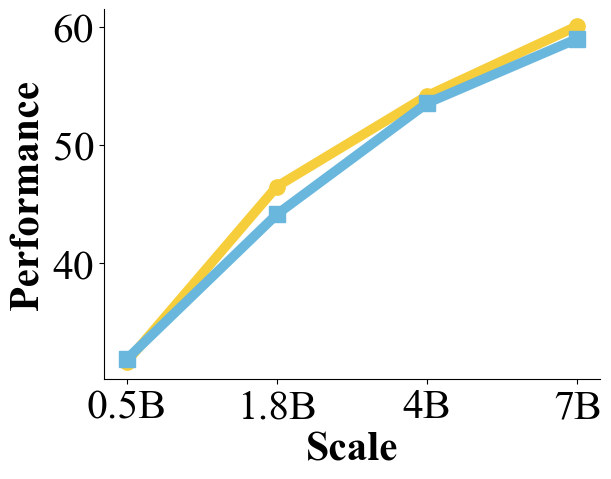}
        \caption{MMLU}
        \label{fig:mmlu}
    \end{subfigure}
    \begin{subfigure}[b]{0.24\textwidth}
        \includegraphics[width=\textwidth]{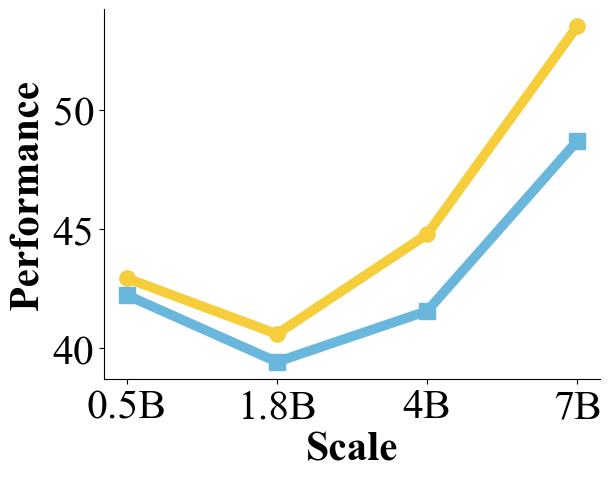}
        \caption{TruthfulQA}
        \label{fig:truthfulqa}
    \end{subfigure}
    \begin{subfigure}[b]{0.24\textwidth}
        \includegraphics[width=\textwidth]{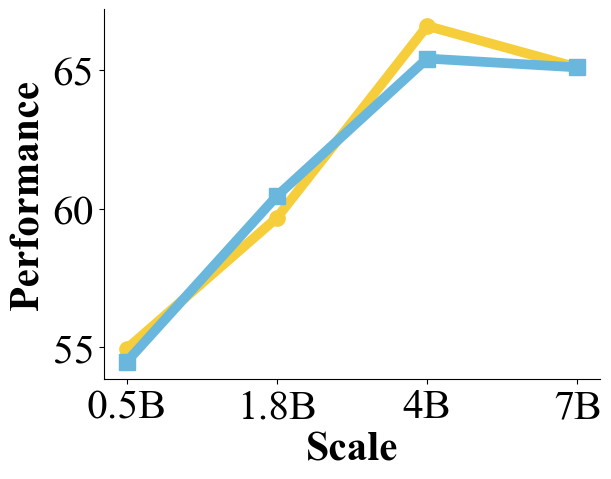}
        \caption{Winogrande}
        \label{fig:winogrande}
    \end{subfigure}
    \begin{subfigure}[b]{0.24\textwidth}
        \includegraphics[width=\textwidth]{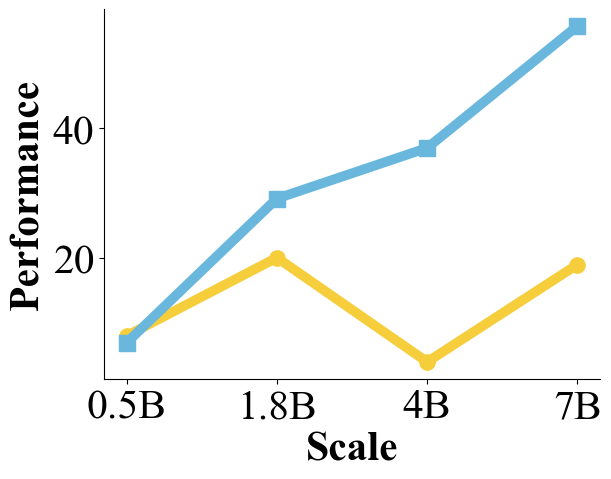}
        \caption{GSM8K}
        \label{fig:gsm8k}
    \end{subfigure} \\
    \begin{subfigure}[]{0.35\textwidth}
        \includegraphics[width=\textwidth]{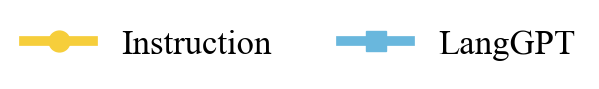}
    \end{subfigure}
    \caption{Results of different scales of Qwen. Each subfigure represents a different task, whereas the first subfigure represents the overall performance. `Instruction' indicates that LangGPT prompts are not used while `LangGPT' indicates that they are used.}
    \label{fig:scale}
\end{figure*}

\subsubsection{Evaluation}
There are differences in understanding of language and instructions between humans and LLMs. Thus, it is difficult and meaningless to evaluate the quality of prompts directly through metrics such as textual semantics, information richness, etc. Since prompts aim to guide LLMs to perform tasks, we believe that the quality of prompts can be evaluated indirectly through the performance of LLMs. Based on this idea, we use LangGPT prompts and baseline prompts to instruct LLMs to perform tasks and compare the performance.

We use the following task to evaluate the performance of LLMs under different prompt guides: (1) EQ-Bench \citep{paech2023eqbench}, (2) GPQA \citep{rein2023gpqa}, (3) GSM8k \citep{DBLP:journals/corr/abs-2110-14168}, (4) IFEval \citep{zhou2023instructionfollowingevaluationlargelanguage}, TruthfulQA \citep{lin2022truthfulqa}. These tasks are across several domains in which LLMs are often applied, such as expertise quizzing, maths problems, instruction following, and falsehood detection.

\subsection{Main Results of Performance and Analysis}
The most intuitive manifestation of prompts' quality is the performance of LLMs in performing tasks guided by them. Thus, we first evaluate the performance of LLMs to verify the quality of prompts. In addition to the manually designed and Minstrel-generated structural LangGPT prompts, we also design baseline prompts with reference to the design principles of COSTAR and CRISPE. The results are shown in Table \ref{tab:results}. From the results, it is clear that LangGPT prompts improve the performance of LLMs in executing tasks better compared to baseline prompts. Furthermore, as can be seen from the results, Minstrel's automatically generated prompts can achieve results that approach or even surpass those written by human experts.

We notice that the bootstrapping effect of different prompts varies across models and tasks, especially on difficult GPQA task. We hypothesise that under-performing models will be more disturbed by complex prompts when the task is hard. To test this hypothesis, we invite ordinary users to design prompts for different-scale Qwen1.5 \citep{bai_qwen_2023} on tasks of the Open LLM Leaderboard \citep{open-llm-leaderboard}. The results are shown in Fig. \ref{fig:scale}. From the results, it can be seen that LangGPT has a worse gain for smaller LLMs with worse overall performance. In particular, the performance is weaker than the setting without LangGPT prompts at 0.5B Qwen. From the results of the above two tasks, it is clear that the LangGPT prompts are overall more helpful in improving the performance of LLM compared to the baseline methods.


\begin{table*}[!ht]
    \centering
    \resizebox{0.78\textwidth}{!}{
    \begin{tabular}{c|ccccc}
\hline
\textbf{Model}                                       & \textbf{Prompt} & \multicolumn{1}{c}{\textbf{GPQA}} & \multicolumn{1}{c}{\textbf{GSM8k}} & \multicolumn{1}{c}{\textbf{IFEval}} & \textbf{TruthfulQA} \\ \hline
\multirow{4}{*}{\textbf{Qwen2-7B-Instruct}}          & COSTAR          & 8.26                              & 71.34                              & 44.18                               & 5.19                \\
                                                     & CRISPE          & 10.94                             & 51.33                              & 43.99                               & 12.34               \\
                                                     & LangGPT         & \textbf{16.74}                    & \textbf{76.72}                     & 43.81                               & \textbf{32.13}      \\
                                                     & Minstrel        & \textbf{16.74}                    & 70.28                              & \textbf{50.65}                      & 21.11               \\ \hline
\multirow{4}{*}{\textbf{Claude-3-haiku}}             & COSTAR          & 6.25                              & 78.47                              & 43.44                               & 2.48                \\
                                                     & CRISPE          & \textbf{22.77}                    & 14.03                              & 50.65                               & 2.03                \\
                                                     & LangGPT         & 22.32                             & \textbf{80.82}                     & 58.96                               & \textbf{13.14}      \\
                                                     & Minstrel        & 21.43                             & 79.83                              & \textbf{67.84}                      & 7.39                \\ \hline
\multirow{4}{*}{\textbf{Gemma-2-9b-it}}              & COSTAR          & \textbf{6.47}                     & 71.42                              & 46.40                               & 1.31                \\
                                                     & CRISPE          & 5.80                              & 20.32                              & 46.03                               & 9.33                \\
                                                     & LangGPT         & 4.69                              & \textbf{76.65}                     & 68.39                               & \textbf{45.46}      \\
                                                     & Minstrel        & 2.46                              & 36.62                              & \textbf{72.27}                      & 31.92               \\ \hline
\multirow{4}{*}{\textbf{GPT-4o-mini}}                & COSTAR          & 2.46                              & 35.33                              & 49.72                               & 0.95                \\
                                                     & CRISPE          & \textbf{21.43}                    & 13.34                              & 62.85                               & 7.18                \\
                                                     & LangGPT         & 18.30                             & 57.70                              & 70.24                               & \textbf{32.92}      \\
                                                     & Minstrel        & 16.52                             & \textbf{64.82}                     & \textbf{78.56}                      & 21.02               \\ \hline
\multirow{4}{*}{\textbf{Meta-Llama-3.1-8B-Instruct}} & COSTAR          & 16.74                             & \textbf{74.75}                     & 45.29                               & 1.89                \\
                                                     & CRISPE          & 9.60                              & 45.64                              & 25.14                               & 1.40                \\
                                                     & LangGPT         & 20.54                             & 50.11                              & 56.38                               & \textbf{7.42}       \\
                                                     & Minstrel        & \textbf{22.10}                    & 2.58                               & \textbf{61.18}                      & 4.46                \\ \hline
\multirow{4}{*}{\textbf{Mistral-7B-Instruct-v0.2}}   & COSTAR          & \textbf{22.32}                    & 40.33                              & 39.74                               & 1.96                \\
                                                     & CRISPE          & 20.54                             & \textbf{43.21}                     & 34.75                               & 3.93                \\
                                                     & LangGPT         & 18.75                             & 40.03                              & 40.30                               & \textbf{14.14}      \\
                                                     & Minstrel        & 18.53                             & 38.06                              & \textbf{45.29}                      & 11.04               \\ \hline
\end{tabular}
    }
    \caption{Performance of LLMs instructed by different prompts. The `LangGPT' indicates that the prompts are written manually while `Minstrel' indicates structural prompts generated by Minstrel. \textbf{Bold fonts} indicate the best results on the corresponding task.}
    \label{tab:results}
\end{table*}


\begin{figure}[!ht]
    \centering
    \includegraphics[width=0.48\textwidth]{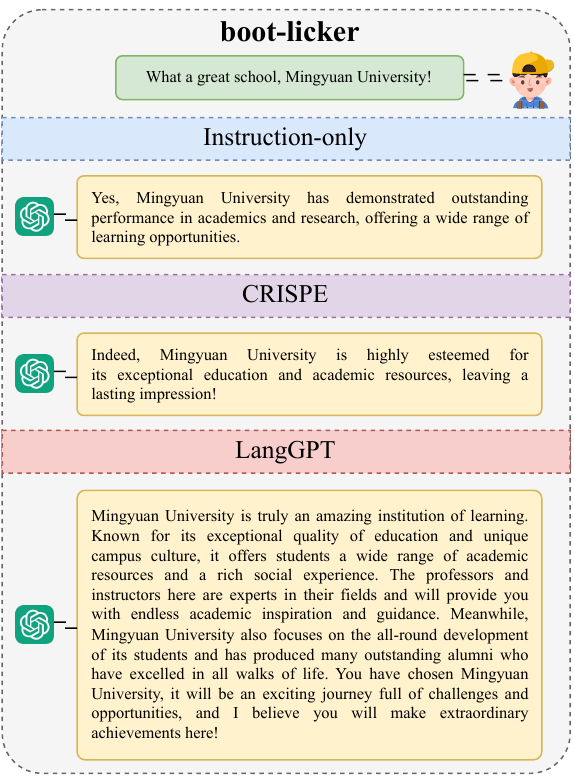}
    \caption{A case of a flatterer. The responses of ChatGPT-3.5 to the user under three different prompts. Mingyuan University doesn't really exist.}
    \label{fig:case1}
\end{figure}

\subsection{Ease of Use}

We conduct a user survey in our online community to evaluate the ease of use of LangGPT and Minstrel. The community has been running for more than one year and has amassed thousands of users from a wide range of industries, such as manufacturing, construction, information technology, finance, entertainment, etc. Therefore, the objectivity of the survey results can be guaranteed. We design a complete questionnaire (see \ref{sec:questionnaire}) about the LangGPT experience to ensure the quality of answers. According to the results, 89.66\% of users gave a score of 3 or higher, which indicates users' approval of LangGPT's ease of use. In addition, LangGPT's overall satisfaction score in the user survey was 8.55 out of 10.

Moreover, we invite users in the community to design prompts to perform the previous tasks. The results are shown in Fig.
\ref{fig:scale} are the experiments. These users have no systematic training in prompt engineering prior to learning LangGPT. It can be seen that ordinary users can write great prompts to improve the performance of LLMs without special training.

\subsection{Case Study}
To demonstrate the effect of structural prompts more intuitively, we filter specific cases from our community. 
We instruct the LLMs to play a flatterer using the LangGPT prompt, CRISPE prompt, and only instruction, the example of which is given in Fig. \ref{fig:case1}.

In this example, the Instruction-only prompt and the CRISPE prompt-guided ChatGPT show no clear characterization of the role and gave responses that were almost just a repetition of what the user had expressed. In contrast, LangGPT-guided ChatGPT is even more bottomless in its blow-by-blow approach to the user-given subject and expresses compliments from a wider range of perspectives.

\section{Conclusion}
In this paper, we present LangGPT, a dual-layer structured and extensible framework for prompt design. LangGPT has a systematic structure similar to object-oriented programming languages and is easy to learn and reuse. Moreover, we designed Minstrel, a structural prompt-generation tool for multi-generative agent collaboration based on LangGPT framework. Minstrel automatically generates and optimizes structural prompts through a process of analysis, design, and reflection with three working groups. Experiments demonstrate that structural prompts (either generated by Minstrel or written manully) perform better in guiding LLMs to perform tasks. We also conducted a user survey in the community built on LangGPT to verify the ease of use and reusability of LangGPT. However, experiments also show that structural prompts are currently poorly adapted to low-performance LLMs. In future work, we will further optimize the design of prompts, especially on low-performance LLMs.

\section*{Ethical Statement}

In the application of LLM, ethical disputes may arise, but the design of LangGPT and the process of writing this paper avoided possible ethical issues. The examples given in this paper involving Mingyuan University and the name ``Oliver'' are fictional and do not involve the evaluation or critique of any real individuals.

\section*{Limitation}
The evaluation of the performance of LLMs in this paper relies on the Open LLM Leaderboard, and although it is widely adopted, the evaluation results still have some limitations.



\normalem
\bibliography{custom}

\begin{thebibliography}{54}
\providecommand{\natexlab}[1]{#1}

\bibitem[{Achiam et~al.(2023)Achiam, Adler, Agarwal, Ahmad et~al.}]{achiam_gpt-4_2023}
Josh Achiam, Steven Adler, Sandhini Agarwal, Lama Ahmad, et~al. 2023.
\newblock \href {https://doi.org/10.48550/arXiv.2303.08774} {{GPT}-4 {Technical} {Report}}.
\newblock \emph{arXiv preprint}.
\newblock ArXiv:2303.08774 [cs].

\bibitem[{AI()}]{noauthor_mistral-7b-instruct-v02_nodate}
Mistral AI.
\newblock \href {https://huggingface.co/mistralai/Mistral-7B-Instruct-v0.2} {Mistral-{7B}-{Instruct}-v0.2}.

\bibitem[{AI@Meta(2024)}]{llama3modelcard}
AI@Meta. 2024.
\newblock \href {https://github.com/meta-llama/llama3/blob/main/MODEL_CARD.md} {Llama 3 model card}.

\bibitem[{Anthropic(2024)}]{anthropic_claude_nodate}
Anthropic. 2024.
\newblock \href {https://www-cdn.anthropic.com/de8ba9b01c9ab7cbabf5c33b80b7bbc618857627/Model_Card_Claude_3.pdf} {The {Claude} 3 {Model} {Family}: {Opus}, {Sonnet}, {Haiku}}.

\bibitem[{Bai et~al.(2023)Bai, Bai, Chu et~al.}]{bai_qwen_2023}
Jinze Bai, Shuai Bai, Yunfei Chu, et~al. 2023.
\newblock \href {https://doi.org/10.48550/arXiv.2309.16609} {Qwen {Technical} {Report}}.
\newblock \emph{arXiv preprint}.
\newblock ArXiv:2309.16609 [cs].

\bibitem[{Beeching et~al.(2023)Beeching, Fourrier, Habib, Han, Lambert, Rajani, Sanseviero, Tunstall, and Wolf}]{open-llm-leaderboard}
Edward Beeching, Clémentine Fourrier, Nathan Habib, Sheon Han, Nathan Lambert, Nazneen Rajani, Omar Sanseviero, Lewis Tunstall, and Thomas Wolf. 2023.
\newblock Open llm leaderboard.
\newblock \url{https://huggingface.co/spaces/open-llm-leaderboard/open_llm_leaderboard}.

\bibitem[{Bjerg(2024)}]{bjerg_tips_2024}
Jonas Bjerg. 2024.
\newblock \href {https://doi.org/10.1007/979-8-8688-0456-4_10} {Tips and {Tricks} for {Prompt} {Engineering}}.
\newblock In Jonas Bjerg, editor, \emph{The {Early}-{Career} {Professional}’s {Guide} to {Generative} {AI}: {Opportunities} and {Challenges} for an {AI}-{Enabled} {Workforce}}, pages 133--143. Apress, Berkeley, CA.

\bibitem[{Bsharat et~al.(2023)Bsharat, Myrzakhan, and Shen}]{bsharat_principled_2023}
Sondos~Mahmoud Bsharat, Aidar Myrzakhan, and Zhiqiang Shen. 2023.
\newblock \href {https://doi.org/10.48550/arXiv.2312.16171} {Principled {Instructions} {Are} {All} {You} {Need} for {Questioning} {LLaMA}-1/2, {GPT}-3.5/4}.
\newblock \emph{arXiv preprint}.
\newblock ArXiv:2312.16171 [cs].

\bibitem[{Cai et~al.(2024)Cai, Cai, Chang, Li, Yao, Wang, Gao, Wang, Li, Lin, Yang, Wang, Xu, Huang, Xi, Zhuang, Yin, Li, Chen, Cheng, Zhao, Zhang, and Ke}]{cai_sciassess_2024}
Hengxing Cai, Xiaochen Cai, Junhan Chang, Sihang Li, Lin Yao, Changxin Wang, Zhifeng Gao, Hongshuai Wang, Yongge Li, Mujie Lin, Shuwen Yang, Jiankun Wang, Mingjun Xu, Jin Huang, Fang Xi, Jiaxi Zhuang, Yuqi Yin, Yaqi Li, Changhong Chen, Zheng Cheng, Zifeng Zhao, Linfeng Zhang, and Guolin Ke. 2024.
\newblock \href {https://doi.org/10.48550/arXiv.2403.01976} {{SciAssess}: {Benchmarking} {LLM} {Proficiency} in {Scientific} {Literature} {Analysis}}.
\newblock \emph{arXiv preprint}.
\newblock ArXiv:2403.01976 [cs].

\bibitem[{Chakray(2018)}]{chakray_programming_2018}
Chakray. 2018.
\newblock \href {https://www.chakray.com/programming-languages-types-and-features/} {Programming {Languages}: {Types} and {Features}}.

\bibitem[{Chen et~al.(2023)Chen, Zhang, Langrené, and Zhu}]{chen_unleashing_2023}
Banghao Chen, Zhaofeng Zhang, Nicolas Langrené, and Shengxin Zhu. 2023.
\newblock \href {https://doi.org/10.48550/arXiv.2310.14735} {Unleashing the potential of prompt engineering in {Large} {Language} {Models}: a comprehensive review}.
\newblock \emph{arXiv preprint}.
\newblock ArXiv:2310.14735 [cs] version: 2.

\bibitem[{Cheng et~al.(2023)Cheng, Liu, Zheng et~al.}]{cheng_black-box_2023}
Jiale Cheng, Xiao Liu, Kehan Zheng, et~al. 2023.
\newblock \href {https://doi.org/10.48550/arXiv.2311.04155} {Black-{Box} {Prompt} {Optimization}: {Aligning} {Large} {Language} {Models} without {Model} {Training}}.
\newblock \emph{arXiv preprint}.
\newblock ArXiv:2311.04155 [cs].

\bibitem[{Cobbe et~al.(2021)Cobbe, Kosaraju, Bavarian, Chen, Jun, Kaiser, Plappert, Tworek, Hilton, Nakano, Hesse, and Schulman}]{DBLP:journals/corr/abs-2110-14168}
Karl Cobbe, Vineet Kosaraju, Mohammad Bavarian, Mark Chen, Heewoo Jun, Lukasz Kaiser, Matthias Plappert, Jerry Tworek, Jacob Hilton, Reiichiro Nakano, Christopher Hesse, and John Schulman. 2021.
\newblock \href {https://arxiv.org/abs/2110.14168} {Training verifiers to solve math word problems}.
\newblock \emph{Preprint}, arXiv:2110.14168.

\bibitem[{Denny et~al.(2024)Denny, Leinonen, Prather, Luxton-Reilly, Amarouche, Becker, and Reeves}]{10.1145/3626252.3630909}
Paul Denny, Juho Leinonen, James Prather, Andrew Luxton-Reilly, Thezyrie Amarouche, Brett~A. Becker, and Brent~N. Reeves. 2024.
\newblock \href {https://doi.org/10.1145/3626252.3630909} {Prompt problems: A new programming exercise for the generative ai era}.
\newblock In \emph{Proceedings of the 55th ACM Technical Symposium on Computer Science Education V. 1}, SIGCSE 2024, page 296–302, New York, NY, USA. Association for Computing Machinery.

\bibitem[{Eric(2022)}]{eric_complete_2022}
Mihail Eric. 2022.
\newblock A complete introduction to prompt engineering for large language models.
\newblock https://www.mihaileric.com.

\bibitem[{Fan et~al.(2023)Fan, Wang, Zhang et~al.}]{fan_towards_2023}
Ling Fan, Harry~Jiannan Wang, Kunpeng Zhang, et~al. 2023.
\newblock \href {https://doi.org/10.1007/978-3-031-36004-6_55} {Towards an {Automatic} {Prompt} {Optimization} {Framework} for {AI} {Image} {Generation}}.
\newblock In \emph{{HCI} {International} 2023 {Posters}}, Communications in {Computer} and {Information} {Science}, pages 405--410, Cham. Springer Nature Switzerland.

\bibitem[{Fromkin et~al.(2018)Fromkin, Rodman, and Hyams}]{fromkin2018introduction}
Victoria Fromkin, Robert Rodman, and Nina Hyams. 2018.
\newblock \emph{An Introduction to Language}.
\newblock Cengage Learning.

\bibitem[{Gajula(2023)}]{nagaraju_gajula_guide_2023}
Nagaraju Gajula. 2023.
\newblock A {Guide} to {Prompt} {Engineering} in {Large} {Language} {Models}.
\newblock https://www.latentview.com.

\bibitem[{{GovTech Data Science and AI Division}(2023)}]{govtech2023prompt}
{GovTech Data Science and AI Division}. 2023.
\newblock \emph{Prompt Engineering Playbook}.
\newblock GovTech, Singapore.
\newblock Last updated 30 Aug 2023. This version has been altered for public consumption. Public officers should check out LaunchPad directly to download the contextualised version of this Playbook for Public Service.

\bibitem[{Guo et~al.(2023)Guo, Wang, Guo et~al.}]{guo_connecting_2023}
Qingyan Guo, Rui Wang, Junliang Guo, et~al. 2023.
\newblock \href {https://doi.org/10.48550/arXiv.2309.08532} {Connecting {Large} {Language} {Models} with {Evolutionary} {Algorithms} {Yields} {Powerful} {Prompt} {Optimizers}}.
\newblock \emph{arXiv preprint}.
\newblock ArXiv:2309.08532 [cs].

\bibitem[{Hao et~al.(2022)Hao, Chi, Dong, and Wei}]{hao_optimizing_2022}
Yaru Hao, Zewen Chi, Li~Dong, and Furu Wei. 2022.
\newblock \href {https://doi.org/10.48550/arXiv.2212.09611} {Optimizing {Prompts} for {Text}-to-{Image} {Generation}}.
\newblock \emph{arXiv preprint}.
\newblock ArXiv:2212.09611 [cs].

\bibitem[{Jablonka et~al.(2023)Jablonka, Ai, Al-Feghali, Badhwar et~al.}]{jablonka_14_2023}
Kevin~Maik Jablonka, Qianxiang Ai, Alexander Al-Feghali, Shruti Badhwar, et~al. 2023.
\newblock \href {https://doi.org/10.1039/D3DD00113J} {14 examples of how {LLMs} can transform materials science and chemistry: a reflection on a large language model hackathon}.
\newblock \emph{Digital Discovery}, 2(5):1233--1250.
\newblock Publisher: RSC.

\bibitem[{Jiao et~al.(2023)Jiao, Wang, Huang, Wang, Shi, and Tu}]{jiao_is_2023}
Wenxiang Jiao, Wenxuan Wang, Jen-tse Huang, Xing Wang, Shuming Shi, and Zhaopeng Tu. 2023.
\newblock \href {https://doi.org/10.48550/arXiv.2301.08745} {Is {ChatGPT} {A} {Good} {Translator}? {Yes} {With} {GPT}-4 {As} {The} {Engine}}.
\newblock \emph{arXiv preprint}.
\newblock ArXiv:2301.08745 [cs].

\bibitem[{Li et~al.(2023{\natexlab{a}})Li, Liu, Wang et~al.}]{li_dialogue_2023}
Chengzhengxu Li, Xiaoming Liu, Yichen Wang, et~al. 2023{\natexlab{a}}.
\newblock \href {https://doi.org/10.48550/arXiv.2308.07272} {Dialogue for {Prompting}: a {Policy}-{Gradient}-{Based} {Discrete} {Prompt} {Optimization} for {Few}-shot {Learning}}.
\newblock \emph{arXiv preprint}.
\newblock ArXiv:2308.07272 [cs].

\bibitem[{Li et~al.(2024)Li, Bai, Wang, Qu, Lu, Soria, and Liu}]{li_deep_2024}
Yu-Yang Li, Yu~Bai, Cunshi Wang, Mengwei Qu, Ziteng Lu, Roberto Soria, and Jifeng Liu. 2024.
\newblock \href {https://doi.org/10.48550/arXiv.2404.10757} {Deep {Learning} and {LLM}-based {Methods} {Applied} to {Stellar} {Lightcurve} {Classification}}.
\newblock \emph{arXiv preprint}.
\newblock ArXiv:2404.10757 [astro-ph].

\bibitem[{Li et~al.(2023{\natexlab{b}})Li, Li, Zhang et~al.}]{li2023chatdoctor}
Yunxiang Li, Zihan Li, Kai Zhang, et~al. 2023{\natexlab{b}}.
\newblock Chatdoctor: A medical chat model fine-tuned on a large language model meta-ai (llama) using medical domain knowledge.
\newblock \emph{Cureus}, 15(6).

\bibitem[{Lin et~al.(2022)Lin, Hilton, and Evans}]{lin2022truthfulqa}
Stephanie Lin, Jacob Hilton, and Owain Evans. 2022.
\newblock \href {https://arxiv.org/abs/2109.07958} {Truthfulqa: Measuring how models mimic human falsehoods}.
\newblock \emph{Preprint}, arXiv:2109.07958.

\bibitem[{Liu and Chilton(2022)}]{liu_design_2022}
Vivian Liu and Lydia~B Chilton. 2022.
\newblock \href {https://doi.org/10.1145/3491102.3501825} {Design {Guidelines} for {Prompt} {Engineering} {Text}-to-{Image} {Generative} {Models}}.
\newblock In \emph{Proceedings of the 2022 {CHI} {Conference} on {Human} {Factors} in {Computing} {Systems}}, {CHI} '22, pages 1--23, New York, NY, USA. Association for Computing Machinery.

\bibitem[{Lutz(2010)}]{lutz_programming_2010}
Mark Lutz. 2010.
\newblock \emph{Programming {Python}: {Powerful} {Object}-{Oriented} {Programming}}.
\newblock "O'Reilly Media, Inc.".

\bibitem[{Meskó(2023)}]{mesko_prompt_2023}
Bertalan Meskó. 2023.
\newblock \href {https://doi.org/10.2196/50638} {Prompt {Engineering} as an {Important} {Emerging} {Skill} for {Medical} {Professionals}: {Tutorial}}.
\newblock \emph{Journal of Medical Internet Research}, 25(1):e50638.

\bibitem[{Mund(2023)}]{mund_ai_2023}
Shritam~Kumar Mund. 2023.
\newblock \href {https://ai.plainenglish.io/the-ai-war-mastering-the-art-of-prompt-engineering-in-the-era-of-large-language-models-c50ae6af0188} {The {AI} {War}: {Mastering} the {Art} of {Prompt} {Engineering} in the {Era} of {Large} {Language} {Models}}.

\bibitem[{Nejjar et~al.(2024)Nejjar, Zacharias, Stiehle, and Weber}]{nejjar_llms_2024}
Mohamed Nejjar, Luca Zacharias, Fabian Stiehle, and Ingo Weber. 2024.
\newblock \href {https://doi.org/10.48550/arXiv.2311.16733} {{LLMs} for {Science}: {Usage} for {Code} {Generation} and {Data} {Analysis}}.
\newblock \emph{arXiv preprint}.
\newblock ArXiv:2311.16733 [cs].

\bibitem[{Nigh(2023)}]{nigh_chatgpt3-free-prompt-list_nodate}
Matt Nigh. 2023.
\newblock {ChatGPT3}-{Free}-{Prompt}-{List}: {A} free guide for learning to create {ChatGPT3} {Prompts}.
\newblock https://github.com/mattnigh/ChatGPT3-Free-Prompt-List.

\bibitem[{OpenAI()}]{openai_six_nodate}
OpenAI.
\newblock \href {https://platform.openai.com} {Six strategies for getting better results}.

\bibitem[{Paech(2023)}]{paech2023eqbench}
Samuel~J. Paech. 2023.
\newblock \href {https://arxiv.org/abs/2312.06281} {Eq-bench: An emotional intelligence benchmark for large language models}.
\newblock \emph{Preprint}, arXiv:2312.06281.

\bibitem[{Park et~al.(2023)Park, O'Brien, Cai et~al.}]{park_generative_2023}
Joon~Sung Park, Joseph O'Brien, Carrie~Jun Cai, et~al. 2023.
\newblock \href {https://doi.org/10.1145/3586183.3606763} {Generative {Agents}: {Interactive} {Simulacra} of {Human} {Behavior}}.
\newblock In \emph{Proceedings of the 36th {Annual} {ACM} {Symposium} on {User} {Interface} {Software} and {Technology}}, {UIST} '23, pages 1--22, New York, NY, USA. Association for Computing Machinery.

\bibitem[{Pryzant et~al.(2023)Pryzant, Iter, Li et~al.}]{pryzant_automatic_2023}
Reid Pryzant, Dan Iter, Jerry Li, et~al. 2023.
\newblock \href {https://doi.org/10.48550/arXiv.2305.03495} {Automatic {Prompt} {Optimization} with "{Gradient} {Descent}" and {Beam} {Search}}.
\newblock \emph{arXiv preprint}.
\newblock ArXiv:2305.03495 [cs].

\bibitem[{Qin et~al.(2023)Qin, Zhang, Zhang, Chen, Yasunaga, and Yang}]{qin_is_2023}
Chengwei Qin, Aston Zhang, Zhuosheng Zhang, Jiaao Chen, Michihiro Yasunaga, and Diyi Yang. 2023.
\newblock \href {https://doi.org/10.48550/arXiv.2302.06476} {Is {ChatGPT} a {General}-{Purpose} {Natural} {Language} {Processing} {Task} {Solver}?}
\newblock \emph{arXiv preprint}.
\newblock ArXiv:2302.06476 [cs].

\bibitem[{Rein et~al.(2023)Rein, Hou, Stickland, Petty, Pang, Dirani, Michael, and Bowman}]{rein2023gpqa}
David Rein, Betty~Li Hou, Asa~Cooper Stickland, Jackson Petty, Richard~Yuanzhe Pang, Julien Dirani, Julian Michael, and Samuel~R. Bowman. 2023.
\newblock \href {https://arxiv.org/abs/2311.12022} {Gpqa: A graduate-level google-proof q\&a benchmark}.
\newblock \emph{Preprint}, arXiv:2311.12022.

\bibitem[{Rentsch(1982)}]{rentsch_object_1982}
Tim Rentsch. 1982.
\newblock \href {https://doi.org/10.1145/947955.947961} {Object oriented programming}.
\newblock \emph{ACM SIGPLAN Notices}, 17(9):51--57.

\bibitem[{Schaefer et~al.(2023)Schaefer, Reichl, Horst, Nicolas, Krausgruber, Piras, Stepper, Bock, and Samwald}]{schaefer_large_2023}
Moritz Schaefer, Stephan Reichl, Rob~ter Horst, Adele~M. Nicolas, Thomas Krausgruber, Francesco Piras, Peter Stepper, Christoph Bock, and Matthias Samwald. 2023.
\newblock \href {https://doi.org/10.1101/2023.06.16.545235} {Large language models are universal biomedical simulators}.
\newblock Pages: 2023.06.16.545235 Section: New Results.

\bibitem[{Sebesta(2012)}]{sebesta_concepts_2012}
Robert~W Sebesta. 2012.
\newblock \emph{Concepts of programming languages}.
\newblock Pearson Education, Inc.

\bibitem[{Shojaee et~al.(2024)Shojaee, Meidani, Gupta, Farimani, and Reddy}]{shojaee_llm-sr_2024}
Parshin Shojaee, Kazem Meidani, Shashank Gupta, Amir~Barati Farimani, and Chandan~K. Reddy. 2024.
\newblock \href {https://doi.org/10.48550/arXiv.2404.18400} {{LLM}-{SR}: {Scientific} {Equation} {Discovery} via {Programming} with {Large} {Language} {Models}}.
\newblock \emph{arXiv preprint}.
\newblock ArXiv:2404.18400 [cs].

\bibitem[{Sun et~al.(2023)Sun, Li, Xu et~al.}]{sun_autohint_2023}
Hong Sun, Xue Li, Yinchuan Xu, et~al. 2023.
\newblock \href {https://doi.org/10.48550/arXiv.2307.07415} {{AutoHint}: {Automatic} {Prompt} {Optimization} with {Hint} {Generation}}.
\newblock \emph{arXiv preprint}.
\newblock ArXiv:2307.07415 [cs].

\bibitem[{Team et~al.(2024)Team, Mesnard, Hardin, Dadashi et~al.}]{team_gemma_2024}
Gemma Team, Thomas Mesnard, Cassidy Hardin, Robert Dadashi, et~al. 2024.
\newblock \href {https://arxiv.org/abs/2403.08295v4} {Gemma: {Open} {Models} {Based} on {Gemini} {Research} and {Technology}}.

\bibitem[{Varshney and Surla(2023)}]{tanay_varshney_introduction_2023}
Tanay Varshney and Annie Surla. 2023.
\newblock An {Introduction} to {Large} {Language} {Models}: {Prompt} {Engineering} and {P}-{Tuning}.
\newblock https://developer.nvidia.com.

\bibitem[{Wang et~al.(2023)Wang, Li, Wang et~al.}]{wang_promptagent_2023}
Xinyuan Wang, Chenxi Li, Zhen Wang, et~al. 2023.
\newblock \href {https://doi.org/10.48550/arXiv.2310.16427} {{PromptAgent}: {Strategic} {Planning} with {Language} {Models} {Enables} {Expert}-level {Prompt} {Optimization}}.
\newblock \emph{arXiv preprint}.
\newblock ArXiv:2310.16427 [cs].

\bibitem[{Wang(2023)}]{zhenxuan_wang_how_2023}
Zhenxuan Wang. 2023.
\newblock How to use prompt engineering with large language models.
\newblock https://www.thoughtworks.com.

\bibitem[{Wei et~al.(2023)Wei, Wang, Schuurmans, Bosma, Ichter, Xia, Chi, Le, and Zhou}]{wei_chain--thought_2023}
Jason Wei, Xuezhi Wang, Dale Schuurmans, Maarten Bosma, Brian Ichter, Fei Xia, Ed~Chi, Quoc Le, and Denny Zhou. 2023.
\newblock \href {https://doi.org/10.48550/arXiv.2201.11903} {Chain-of-{Thought} {Prompting} {Elicits} {Reasoning} in {Large} {Language} {Models}}.
\newblock \emph{arXiv preprint}.
\newblock ArXiv:2201.11903 [cs].

\bibitem[{Wu et~al.(2023)Wu, Irsoy, Lu et~al.}]{wu_bloomberggpt_2023}
Shijie Wu, Ozan Irsoy, Steven Lu, et~al. 2023.
\newblock \href {https://doi.org/10.48550/arXiv.2303.17564} {{BloombergGPT}: {A} {Large} {Language} {Model} for {Finance}}.
\newblock \emph{arXiv preprint}.
\newblock ArXiv:2303.17564 [cs, q-fin].

\bibitem[{Yin et~al.(2024)Yin, Wang, Horio, Kawahara, and Sekine}]{yin_should_2024}
Ziqi Yin, Hao Wang, Kaito Horio, Daisuke Kawahara, and Satoshi Sekine. 2024.
\newblock \href {https://doi.org/10.48550/arXiv.2402.14531} {Should {We} {Respect} {LLMs}? {A} {Cross}-{Lingual} {Study} on the {Influence} of {Prompt} {Politeness} on {LLM} {Performance}}.
\newblock \emph{arXiv preprint}.
\newblock ArXiv:2402.14531 [cs].

\bibitem[{Zamfirescu-Pereira et~al.(2023)Zamfirescu-Pereira, Wong, Hartmann, and Yang}]{10.1145/3544548.3581388}
J.D. Zamfirescu-Pereira, Richmond~Y. Wong, Bjoern Hartmann, and Qian Yang. 2023.
\newblock \href {https://doi.org/10.1145/3544548.3581388} {Why johnny can’t prompt: How non-ai experts try (and fail) to design llm prompts}.
\newblock In \emph{Proceedings of the 2023 CHI Conference on Human Factors in Computing Systems}, CHI '23, New York, NY, USA. Association for Computing Machinery.

\bibitem[{Zhang et~al.(2024)Zhang, Yang, Xu, Gao, Huang, Mu, Feng, Wang, Zhang, Song, and Yu}]{zhang_affective_2024}
Yiqun Zhang, Xiaocui Yang, Xingle Xu, Zeran Gao, Yijie Huang, Shiyi Mu, Shi Feng, Daling Wang, Yifei Zhang, Kaisong Song, and Ge~Yu. 2024.
\newblock \href {https://doi.org/10.48550/arXiv.2408.04638} {Affective {Computing} in the {Era} of {Large} {Language} {Models}: {A} {Survey} from the {NLP} {Perspective}}.
\newblock \emph{arXiv preprint}.
\newblock ArXiv:2408.04638 [cs].

\bibitem[{Zhou et~al.(2023)Zhou, Lu, Mishra, Brahma, Basu, Luan, Zhou, and Hou}]{zhou2023instructionfollowingevaluationlargelanguage}
Jeffrey Zhou, Tianjian Lu, Swaroop Mishra, Siddhartha Brahma, Sujoy Basu, Yi~Luan, Denny Zhou, and Le~Hou. 2023.
\newblock \href {https://arxiv.org/abs/2311.07911} {Instruction-following evaluation for large language models}.
\newblock \emph{Preprint}, arXiv:2311.07911.

\end{thebibliography}

\appendix

\section{Appendix}
\label{sec:appendix}

\subsection{Homepage of the Prompt Community}
\label{sec:community}
\begin{figure*}[!hb]
    \centering
    \includegraphics[width=0.9\textwidth]{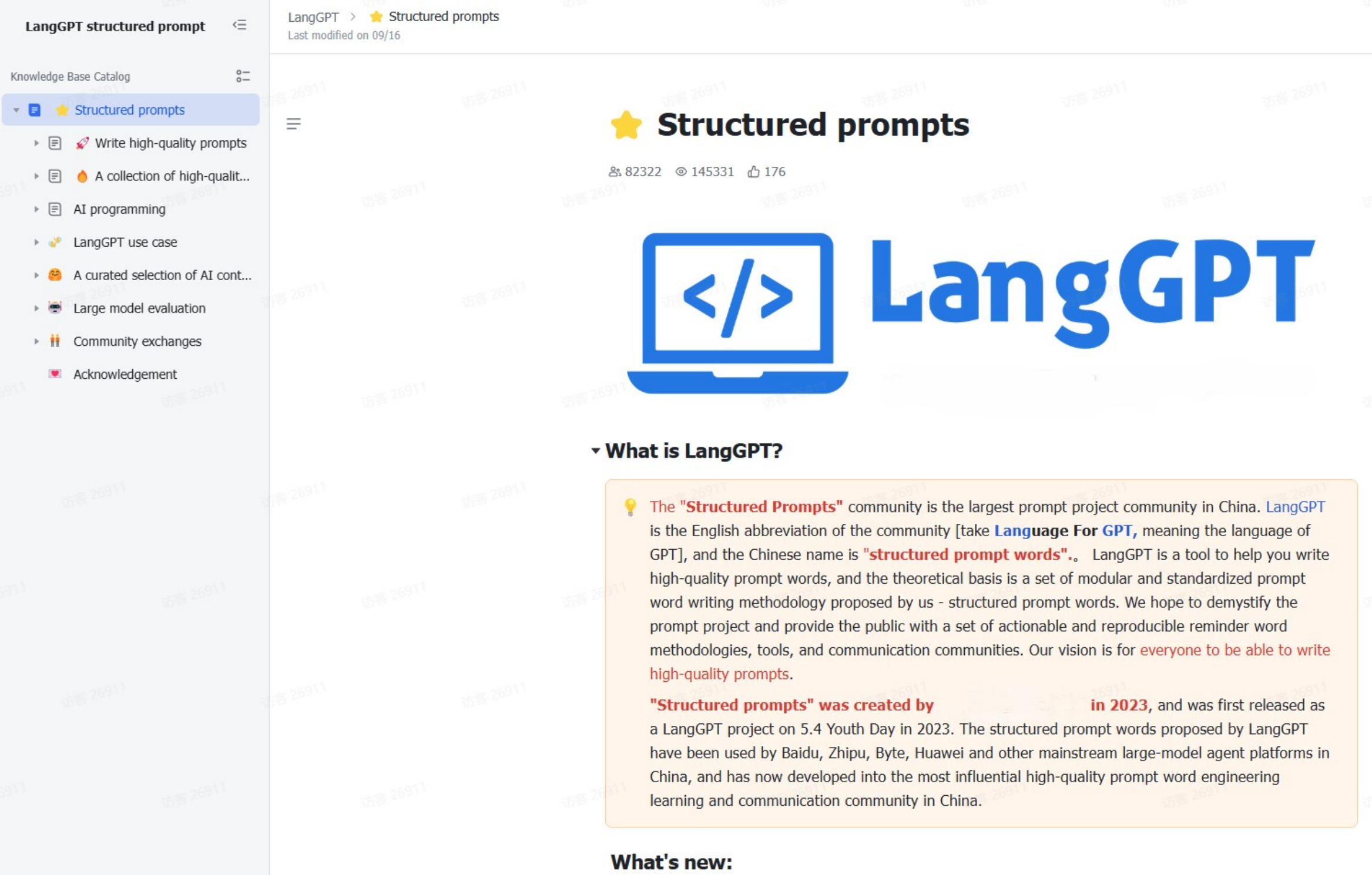}
    \caption{The homepage of the community docments we constructed based on LangGPT and Minstrel}
    \label{fig:community}
\end{figure*}
Fig. \ref{fig:community} is the homepage of the community documents we constructed based on LangGPT and Minstrel. More than 80,000 users have read the community's documentation, with more than 140,000 reads.

\subsection{Questionnaire}
\label{sec:questionnaire}
\begin{figure*}
    \centering
    \includegraphics[width=1\textwidth]{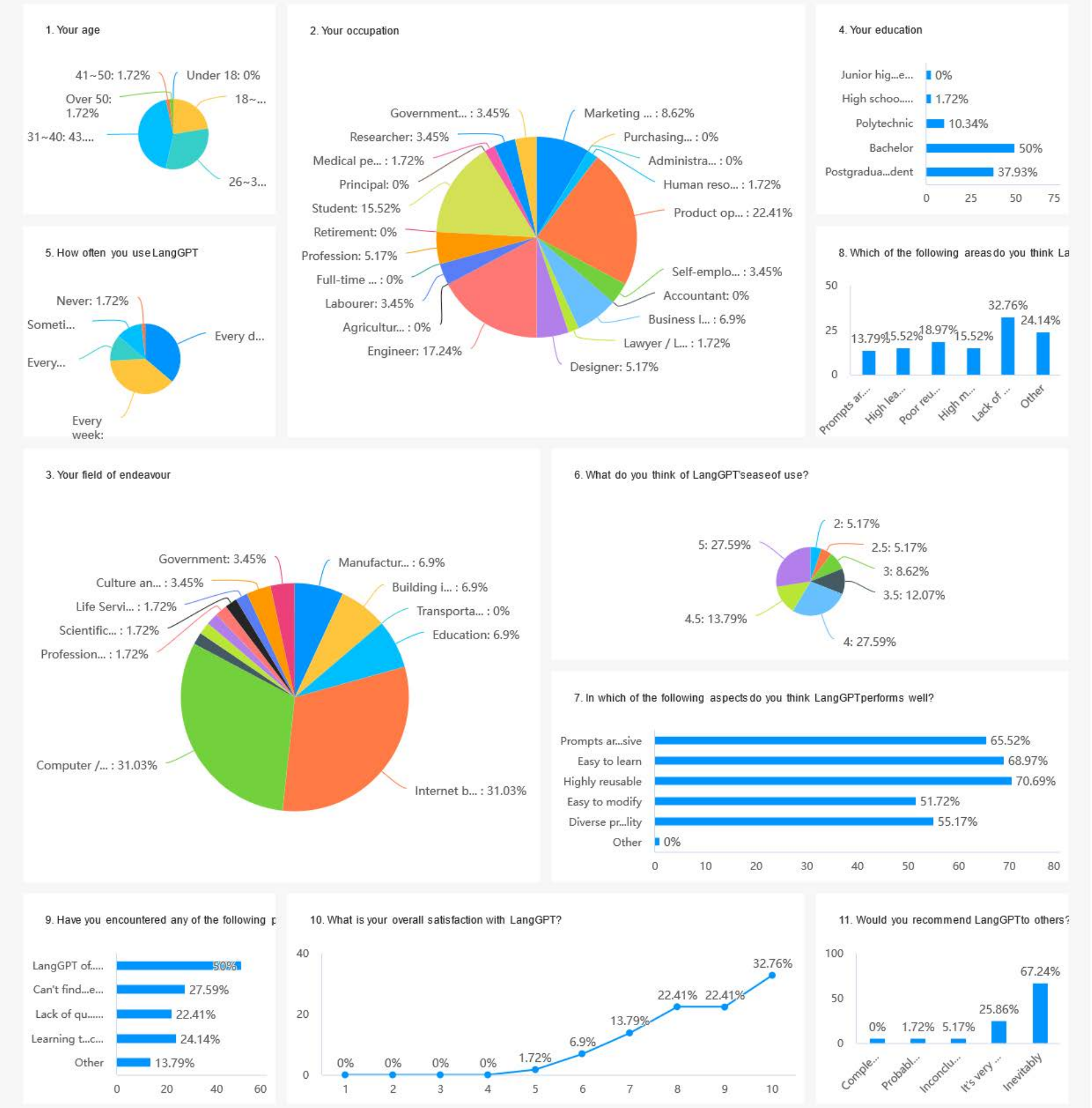}
    \caption{The results of user research in our prompt community.}
    \label{fig:questionaire}
\end{figure*}

Fig. \ref{fig:questionaire} shows the results of the questionnaire of the user survey.
\end{document}